\documentclass[runningheads,]{llncs}
\usepackage{graphicx}
\usepackage{times}
\usepackage{latexsym}
\usepackage[T1]{fontenc}
\usepackage[utf8]{inputenc}

\usepackage{natbib}

\usepackage{url}
\usepackage{graphicx}
\usepackage{multicol}
\usepackage{subfig}

\begin{document}
\title{Interpretable Segmentation  of Medical Free-Text Records Based on Word Embeddings}

%
%
\author{Adam Gabriel Dobrakowski\inst{1}
Agnieszka Mykowiecka\inst{2}
\and
Małgorzata Marciniak\inst{2}
\and
Wojciech Jaworski\inst{1}
\and
Przemys{\l}aw Biecek\inst{1}
}

%
%
\authorrunning{A. Dobrakowski et al.} 
%
\institute{
University of Warsaw, Warsaw, Banacha 2, Poland\\
\email{\{ad359226@students, W.Jaworski, P.Biecek\}@mimuw.edu.pl}
\and
Institute of Computer Science Polish Academy of Sciences, \\Jana Kazimierza 5, Warsaw, Poland\\
\email{\{agn,mm\}@ipipan.waw.pl}
}
\maketitle              

\begin{abstract}
Is it true that patients with similar conditions get similar diagnoses? In this paper we present a natural language processing (NLP) method that can be used to validate this claim.
We (1) introduce a method for representation of medical visits based on free-text descriptions recorded by doctors, (2) introduce a new method for segmentation of patients' visits, (3) present an application of the proposed method on a corpus of 100,000 medical visits and (4) show tools for interpretation and exploration of derived knowledge representation. With the proposed method we obtained stable and separated segments of visits which were positively validated against medical diagnoses. We show how the presented algorithm may be used to aid doctors in their practice.
\end{abstract}

\section{Introduction}

Processing of free-text clinical records plays an~important role in computer-supported medicine \citep{5334831,7004390}. A detailed description of symptoms, examination and an interview is often stored in an unstructured way as free-text, hard to process but rich in important information. Although there exist some attempts to process medical notes for English and some other languages, in general, the problem is still challenging \citep{hungariansegmentation}.
The most straightforward approach to the processing of clinical notes could be their clustering with respect to different features like diagnosis or type of treatment. The process can either concentrate on patients or on their particular visits. 

Grouping of visits can fulfil  many potential goals.  If we are able to group visits into clusters based on interview with a patient and medical examination then we can: follow recommendations that were suggested to patients with similar history to create a list of possible diagnoses;  reveal that the current diagnosis is unusual; identify subsets of visits with the same diagnosis but different symptoms.
A desired goal in the patients' segmentation is to divide them into groups with similar properties. In the case of segmentation hospitalized patients one of the most well-known examples are Diagnosis Related Groups \citep{casemix} which aim to divide patients into groups with similar costs of treatment. Grouping visits of patients in health centers is a different issue. Here most of the information is unstructured and included in the visit's description written by a~doctor: the description of the interview with the patient and the description of a medical examination of the patient.

Segmentation (clustering) is a well studied task  for structured data such as age, sex, place, history of diseases, ICD-10 code etc. (an example of patients segmentation based only on their history of diseases is introduced in \citep{ruffini2017clustering}), but it is far from being solved for unstructured free-texts which requires undertaking many decision on how the text and its meaning is to be represented. 
 Medical concepts to be extracted from texts very often are taken from Unified Medical Language System (UMLS, see \citep{bodenreider2004unified}), which is a commonly accepted base of biomedical terminology. Representations of medical concepts are computed based on various medical texts, like medical journals, books, etc.~\citep{minarro2014exploring,de2014medical,newman2017insights,choi2016learning,chiu2016train} or based directly on data from Electronic Health Records~\citep{choi2016multi,choi2016medical,choi2016learning}.
Other approach for patient segmentation is given in \cite{choi2016multi}. A~subset of medical concepts (e.g. diagnosis, medication, procedures) and  embeddings is aggregated for all visits of a patient. This way we get patient embedding that summaries patient medical history.

In this work we present a different approach. Our data include medical records for the medical history, description of the examination and recommendations for the treatment. Complementary sources allow us to create a more comprehensive visit description. The second difference is grouping visits, not patients. In this way a single patient can belong to several clusters.
Our segmentation is based on a dictionary of medical concepts created from data,  as for Polish does not exist any classification of medical concepts like UMLS or SNOMED.
Obtained segments are supplemented with several approaches to visual exploration that facilitate interpretation of segments. Some examples of visual exploration of supervised models for structured medical data are presented in \cite{Gordon, Kobylinska, DALEX}. In this article we deal with a problem of explainable machine learning for unsupervised models.

\section{Corpus of Free-text Clinical Records}

The clustering method is developed and validated on a  dataset of free-text clinical records of about 100,000 visits.
The data set consists of descriptions of patients' visits from different primary health care centers and specialist clinics in Poland. They have a free-text form and  are written by doctors representing a wide range of medical professions, e.g. general practitioners, dermatologists, cardiologists or psychiatrists. Each description is divided into three parts: interview, examination, and recommendations.

\section{Methodology}

In this section we describe our algorithm for visits clustering. The process is  performed in the following four steps:
(1) Medical concepts are extracted from free-text descriptions of an interview and examination.  (2) A new representation of identified concepts is derived with concepts embedding. (3) Concept embeddings are transformed into visit embeddings.  (4) Clustering is performed on visit embeddings.

\subsection{Extraction of Medical Concepts}

As there are no generally available terminological resources for Polish medical texts, the first step of data processing was aimed at automatic identification of the most frequently used words and phrases. The doctors' notes are usually rather short and concise, so we assumed that all frequently appearing phrases are domain related and important for text understanding. The notes are built mostly from noun phrases which consist of a noun optionally modified by a sequence of adjectives or by another noun in the genitive. We only extracted sequences that can be interpreted as phrases in Polish.

To get the most common phrases, we processed 220,000 visits' descriptions. First, we preprocessed texts using Concraft tagger \citep{wasz:12} which assigns lemmas, POS and morphological features values. It also guesses descriptions (apart from lemmas) for words which  are not present in its vocabulary.  Phrase extraction and ordering was performed by TermoPL \citep{mar:myk:rych:lrec16}. The program  allows for defining a grammar describing extracted text fragments and order them according to a version of the C-value coefficient \citep{fran:etal}, but we used the built-in grammar of  noun phrases. 
 The first 4800 phrases (all with C-value equal at least 20) from the obtained list were manually annotated with semantic labels. The list of 137 labels covered most general concepts like \emph{anatomy, feature, disease, test}.
 Many labels were assigned to multi-word expressions (MWEs). In some cases phrases were also labeled separately, e.g. \emph{left hand} is labeled as \emph{anatomy} while \emph{hand} is also labeled as \emph{anatomy} and \emph{left} as \emph{lateralization}. The additional source of information was the list of 9993 names of medicines and dietary supplements.

The list of terms together with their semantic labels
was then converted to the format of lexical resources of 
Categorial Syntactic-Semantic Parser „ENIAM” \citep{jaworski2016eniam,11321/538}. 
The parser recognized lexemes and MWEs in texts according to the provided list of terms, then the longest sequence of recognized terms was selected, and semantic representation was created. Semantic representation of a visit has a form of a set of pairs composed of recognized terms and their labels (not recognized tokens were omitted). The average coverage of semantic representation was
82.06\% of tokens and 75.38\% of symbols in section \emph{Interview} and 87.43\% of tokens and 79.28\% of symbols in section \emph{Examination}.

Texts of visits are heterogeneous as they consist of: very frequent domain phrases; domain important words which are too infrequent to be at the top of the term list prepared by TermoPL; some general words which do not carry relevant information; numerical information; and words which are misspelled. In the clustering task we neglect the original text with inflected word forms and the experiments are solely performed on the set of semantic labels attached to each interview and examination.  

\subsection{Embeddings for Medical Concepts}

Operating on relatively large amount of very specific texts, we decided not to use any general model for Polish.
In the experiments, we reduce the description of visits to extracted concepts and train on them our own domain embeddings. 
During creating the term co-occurrence matrix the whole visit’s description is treated as the neighbourhood of the concept. Furthermore we choose only unique concepts and abandon their original order in the description (we follow this way due to simplicity).

We compute embeddings of concepts by GloVe \citep{pennington2014glove} for interview descriptions and for examination descriptions separately. Computing two separate embeddings we aim at catching the similarity between terms in their specific context. 
For example,  the nearest words to \textit{cough} in the interview descriptions is \textit{runny nose}, \textit{sore throat}, \textit{fever} but in the examination description it is \textit{rash}, \textit{sunny}, \textit{laryngeal}.

\begin{figure*}[t!]

\centering
\subfloat[Body part -- Right side]{
\includegraphics[width=0.48\linewidth]{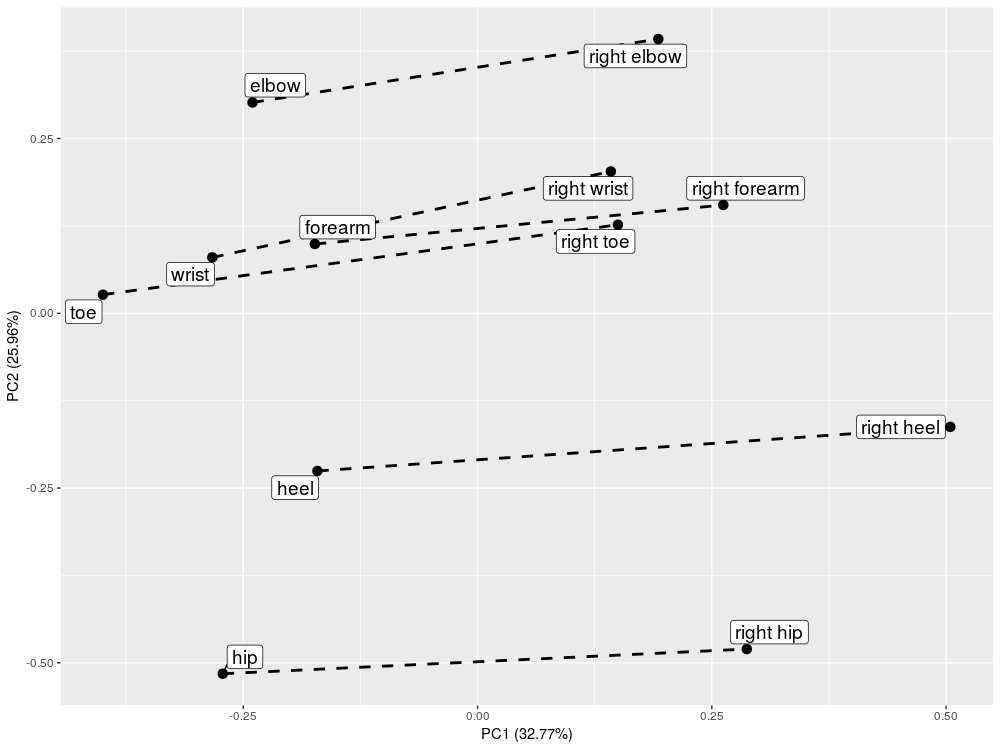}}
\subfloat[Body part -- Pain]{
\includegraphics[width=0.48\linewidth]{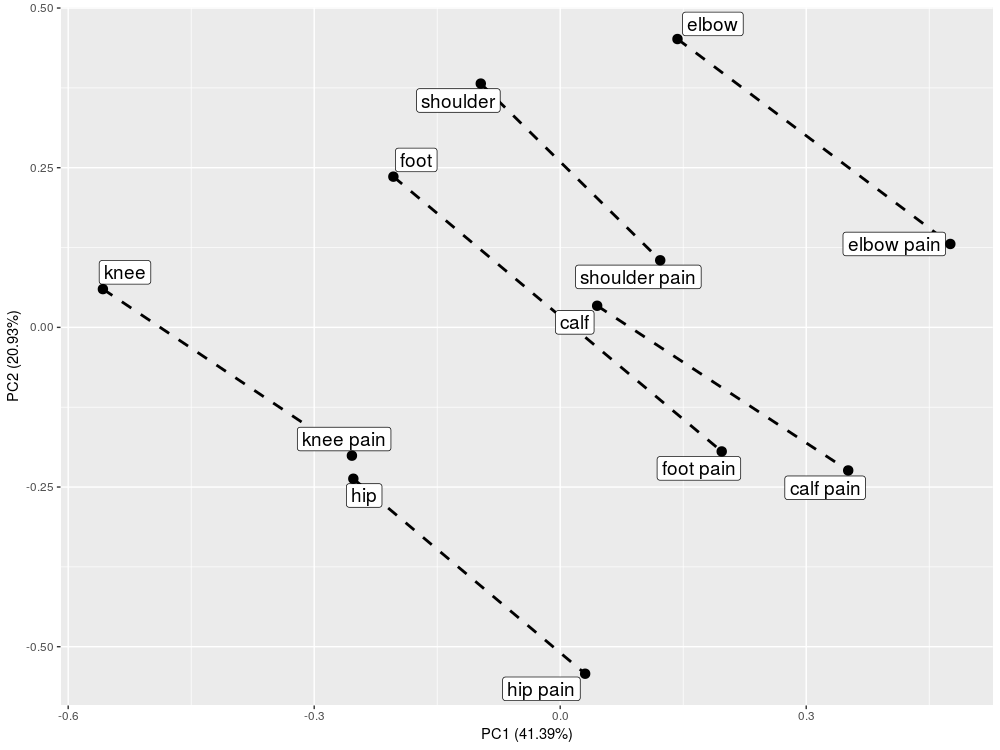}}

\subfloat[Specialty -- Adjective]{
\includegraphics[width=0.48\linewidth]{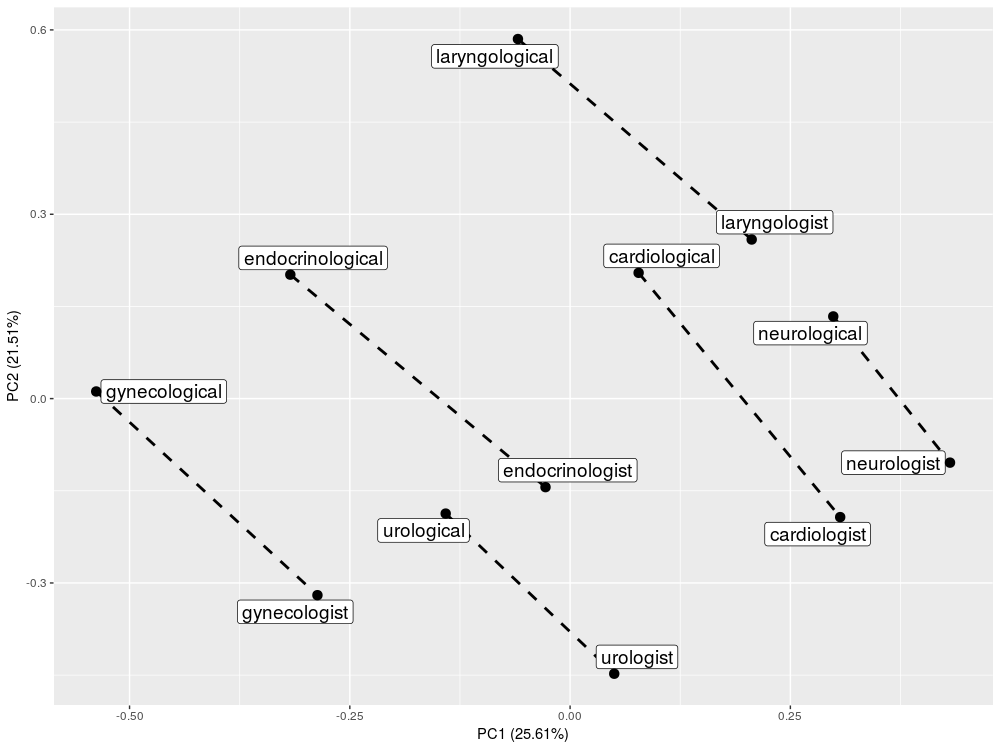}}
\subfloat[Specialty -- Body part]{
\includegraphics[width=0.48\linewidth]{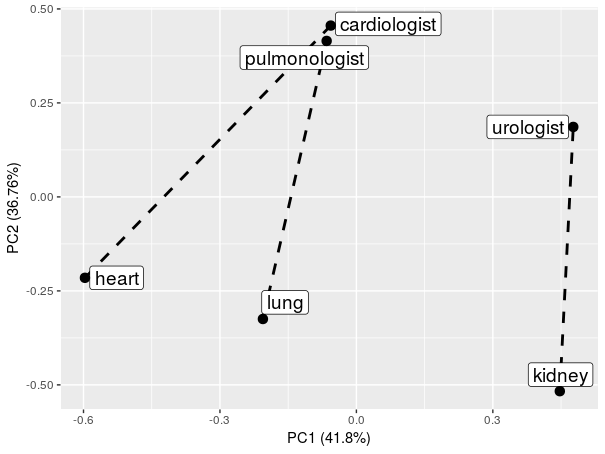}}

\caption{Visualization of analogies between terms. The pictures show term embeddings projected into 2d-plane using PCA. Each panel shows a different type of analogy.}
\label{fig:analogies}
\end{figure*}

\subsection{Visit Embeddings}

The simplest way to generate text embeddings based on term embeddings is to use some kind of aggregation of term embeddings such as an average. This approach was tested for example in~\cite{banea2014simcompass} and \cite{choi2016medical}. In \cite{de2016representation} the authors computed a weighted mean of term embeddings by the construction of a loss function and training weights by the gradient descent method.
Thus, in our method we firstly compute embeddings of the descriptions (for interview and examination separately) as a simple average of concepts' embeddings. Then, the final embeddings for visits are obtained by concatenation of two descriptions' embeddings. 

\subsection{Visits Clustering}

Based on Euclidean distance between vector representations of visits we applied and compared two clustering algorithms: k-means and hierarchical clustering with Ward's method for merging clusters~\citep{ward1963hierarchical}. The similarity of these clusterings was measured by the adjusted Rand index~\citep{rand1971objective}. For the final results we chose the hierarchical clustering algorithm due to greater stability.

\begin{table*}[h!]
\centering
\begin{tabular}{|l||c|c|l|c|} \hline
Domain  & \# clusters & \# visits & clusters' size & K-means \\
& & & & - hclust \\ \hline
 
Cardiology & 6 & 1201 & 428, 193, 134, 303, 27, 116 & 0.87 \\
Family medicine & 6 & 11230 & 3108, 2353, 601, 4518, 255, 395 & 0.69 \\
Gynecology & 4 & 3456 & 1311, 1318, 384, 443 & 0.8 \\
Internal medicine & 5 & 6419 & 1915, 1173, 1930, 1146, 255 & 0.76 \\
Psychiatry & 5 & 1012 & 441, 184, 179, 133, 75 & 0.81  \\
\hline
\end{tabular}
\caption{The statistics of clusters for selected domains. The last column shows adjusted Rand index between k-means and hierarchical clustering.}\label{tab:clusters}

\centering
\begin{tabular}{|l|c||l|l||l|l|} \hline
 Type of relationship & \# Pairs  &
 \multicolumn{2}{c||}{Term Pair 1} & \multicolumn{2}{c|}{Term Pair 2} \\ \hline

Body part -- Pain & 22 & eye & eye pain & foot & foot pain \\
Specialty -- Adjective & 7 & dermatologist & dermatological & neurologist & neurological \\
Body part -- Right side & 34 & hand & right hand & knee & right knee \\
Body part -- Left side & 32 & thumb & left thumb & heel & left heel \\
Spec. -- Consultation & 11 & surgeon & surgical consult. & gynecologist &
g. consult. \\
Specialty - Body part & 9 & cardiologist & heart & oculist & eye \\
Man - Woman & 9 & patient (male) & patient (female) & brother & sister \\ \hline

\end{tabular}
\caption{The categories of questions in term analogy task with example pairs.}\label{tab:analogy_task}

\centering
\begin{tabular}{|c||c|c|c|} \hline
Dim. / Context & 1 & 3 & 5  \\ \hline
 
10 & 0.1293 & 0.2189 & 0.2827 \\
15 & 0.1701 & 0.3081 & 0.4123 \\
20 & \textbf{0.1702} & 0.3749 & 0.4662 \\
25 & 0.1667 & 0.4120 & 0.5220 \\
30 & 0.1674 & 0.4675 & 0.5755 \\
40 & 0.1460 & \textbf{0.5017} & 0.6070 \\
50 & 0.1518 & 0.4966 & \textbf{0.6190} \\
100 & 0.0435 & 0.4231 & 0.5483 \\ 
200 & 0.0261 & 0.3058 & 0.4410 \\ 
\hline
\end{tabular}
\caption{Mean accuracy of correct answers on term analogy tasks.  Rows show different embeddings sizes, columns correspond to size of neighborhoods.}\label{tab:analogies_acc}

    \centering
    \begin{tabular}{c|c|l}
\hline
cluster & size & most frequent recommendations\\
\hline
1 & 1311 & recommendation (16.6\%), general urine test (5.6\%), diet (4.1\%), \\ & & vitamin (4\%), dental prophylaxis (3.4\%)\\
\hline
2 & 1318 & therapy (4.1\%), to treat (4\%), cytology (3.8\%), breast ultrasound (3\%), medicine (2.2\%)\\
\hline
3 & 384 & acidum (31.5\%), the nearest hospital (14.3\%), proper diet (14.3\%), \\ & & health behavior (14.3\%), obstetric control (10.2\%)\\
\hline
4 & 443 & to treat (2\%), therapy (2\%), vitamin (1.8\%), diet (1.6\%), medicine (1.6\%)\\
\hline

    \end{tabular}
    
    \caption{The most common recommendations for each segment derived for gynecologyy. In brackets we present a percentage of visits in this cluster which contain a specified term. We skipped terms common in many clusters, like: \textit{treatment, ultrasound treatment, control, morphology, hospital, lifestyle, zus (Social Insurance Institution)}.}
    \label{tab:recom}

\end{table*}

For clustering, we selected  visits where the description of recommendation and at least one of interview and examination were not empty (some concepts were recognized). It significantly reduced the number of considered visits.
Table~\ref{tab:clusters} gives basic
statistics of obtained clusters. The last column contains the adjusted Rand index. It can be interpreted as a measure of the stability of the clustering. The higher similarity of the two algorithms, the higher stability of clustering.
For determining the optimal number of clusters, for each specialty we consider the number of clusters between 2 and 15. We choose the number of clusters so that adding another cluster does not give a~relevant improvement of a sum of differences between elements and clusters' centers (according to so called
\textit{Elbow method}).

\section{Results}


\subsection{Analogies in Medical Concepts}


To better understand the structure of concept embeddings and to determine the optimal dimension of embedded vectors we use word analogy task introduced in~\cite{mikolov2013efficient} and examined in details in a medical context in~\cite{newman2017insights}. In the former work the authors defined five types of semantic and nine types of syntactic relationship.

We propose our own relationships between concepts, more related to the
medical language. 
We exploit the fact that in the corpus we have a lot of multiword concepts and very often the same words are included in different terms.
We would like the embeddings to be able to catch relationships between terms.
A question in the term analogy task is computing a vector: $vector(left \; foot) - vector(foot) + vector(hand)$ and checking if the correct $vector(left \; hand)$ is in the neighborhood (in the metric of cosine of the angle between the vectors) of this resulting vector.

We defined seven types of such semantic questions and computed answers' accuracy in a similar way as in~\cite{mikolov2013efficient}: we created manually the list of similar term pairs and then we formed the list of questions by taking all two-element subsets of the pairs list. Table~\ref{tab:analogy_task} shows the created categories of questions.

We created one additional task, according to the observation that sometimes two different terms are related to the same object. This can be caused for example by the different order of words in the terms, e.g. \textit{left wrist} and \textit{wrist left} (in Polish both options are acceptable). We checked if the embeddings of such words are similar.

We computed term embeddings for terms occurring at least 5 times in the descriptions of the selected visits. The number of chosen terms in interview descriptions was equal to 3816 and in examination descriptions -- 3559. Among these there were 2556 common terms for interview and examination. Embeddings of the size 10 to 200 were evaluated. For every embedding of interview terms there was measured accuracy of every of eight tasks. Table~\ref{tab:analogies_acc} shows the mean of eight task results. The second column presents the results of the most restrictive rule: a~question is assumed to be correctly answered only if the closest term of the vector computed by operations on related terms is the same as the desired answer. The total number of terms in our data set (about 900,000 for interviews) was many times lower than sets examined in~\cite{mikolov2013efficient}. Furthermore, words in medical descriptions can have a different context that we expect. Taking this into account, the accuracy of about 0.17 is very high and better than we expected.
We then checked the closest 3 and 5 words to the computed vector and assumed a correct answer if in this neighbourhood there was the correct vector. In the biggest neighbourhood the majority of embeddings returned accuracy higher than 0.5.

\begin{figure*}[t!]
\centering
\subfloat[Cardiology]{\label{cardiology}
\includegraphics[width=0.32\linewidth]{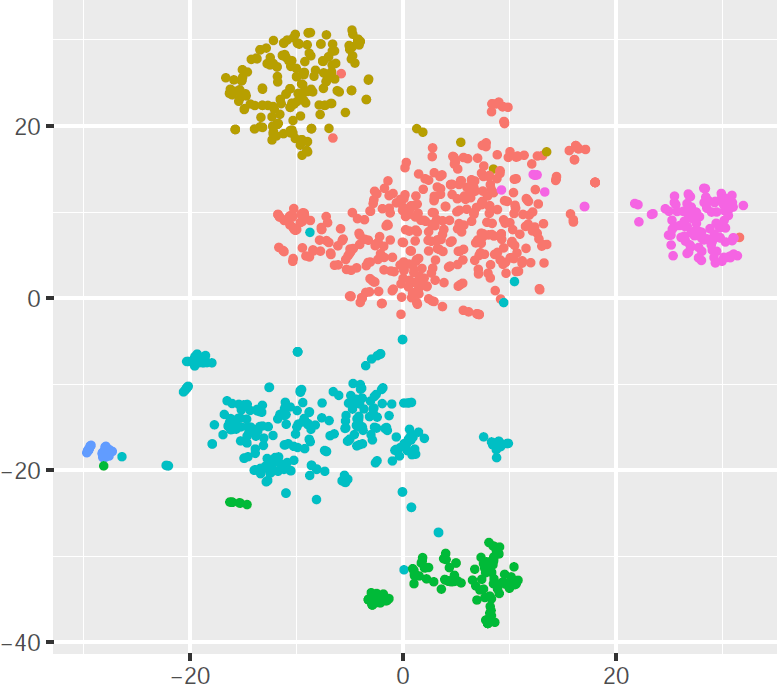}}
\subfloat[Dermatology, venereology]{
\includegraphics[width=0.32\linewidth]{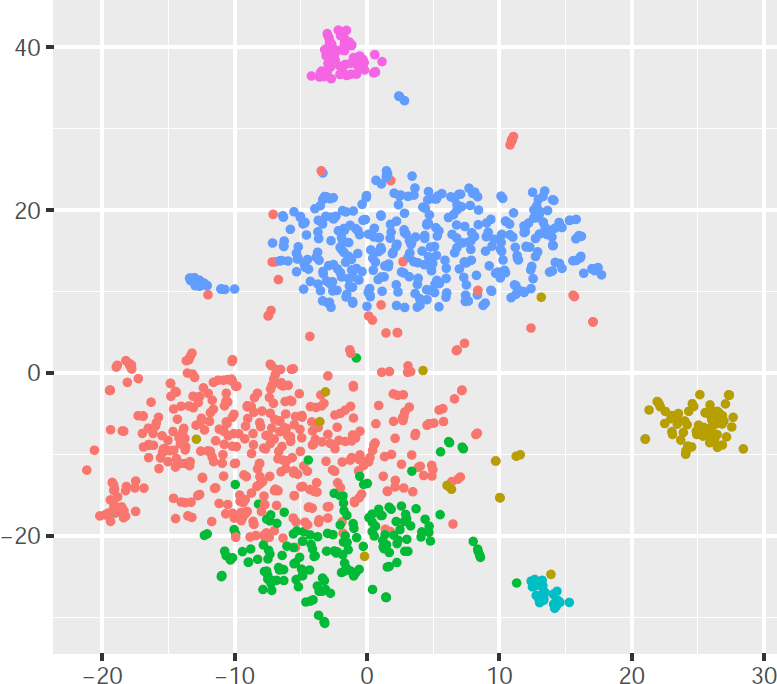}}
\subfloat[Endocrinology]{
\includegraphics[width=0.32\linewidth]{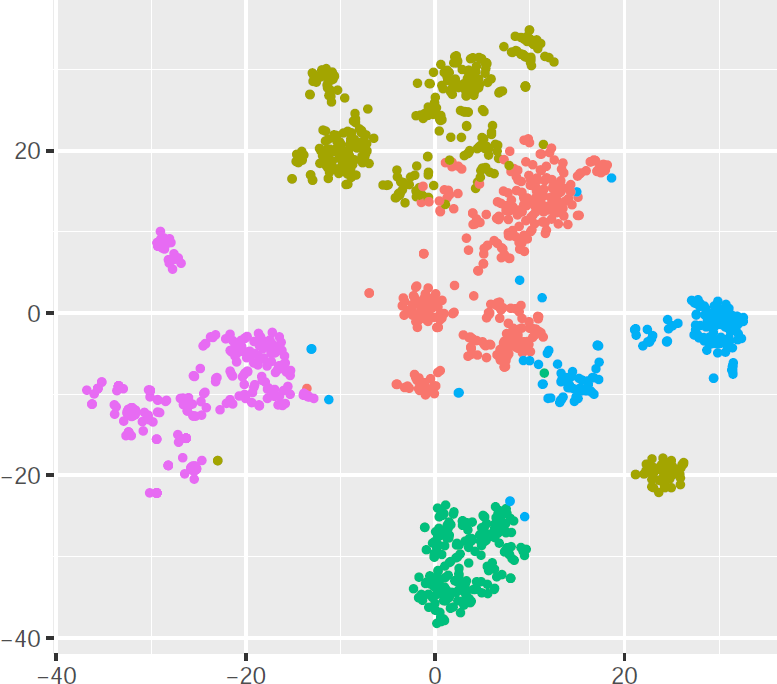}}

\subfloat[Family medicine]{
\includegraphics[width=0.32\linewidth]{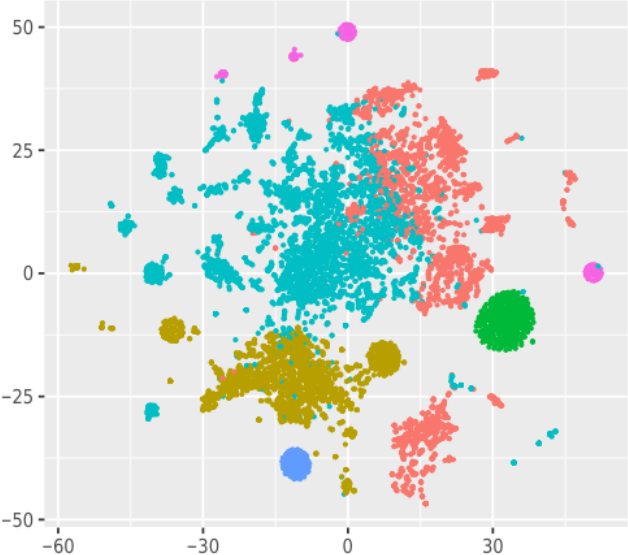}}
\subfloat[Gynecology]{
\includegraphics[width=0.32\linewidth]{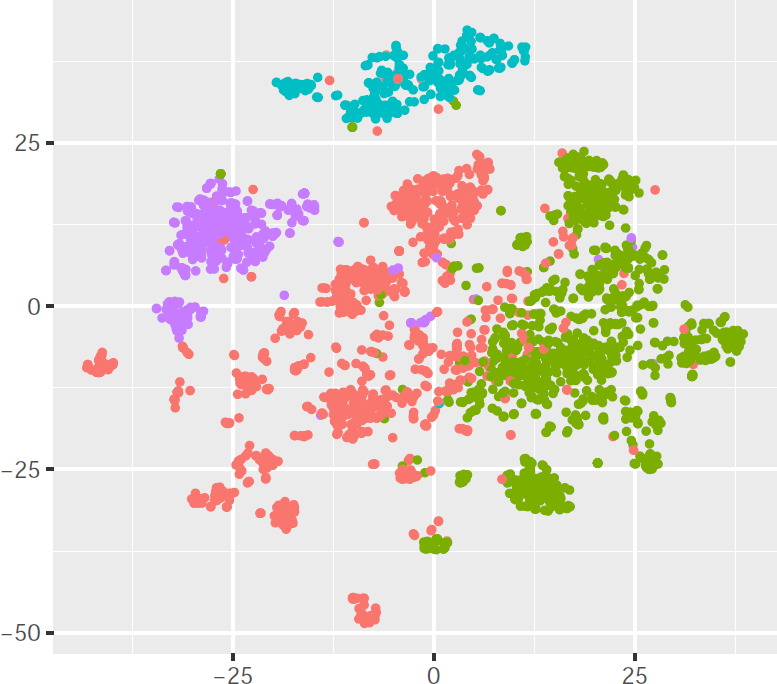}}
\subfloat[Internal medicine]{
\includegraphics[width=0.32\linewidth]{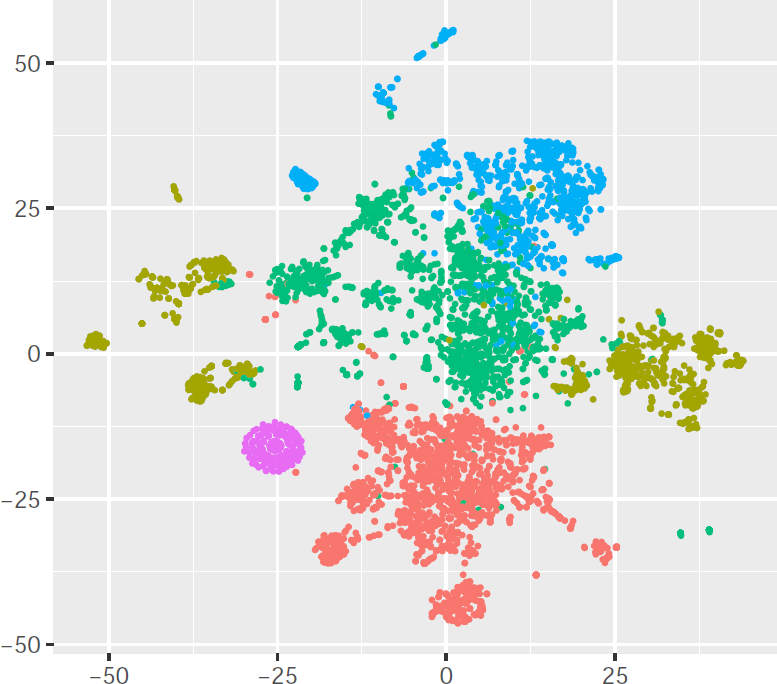}}

\caption{Clusters of visits for selected domains. Each dot corresponds to a single visit. Colors correspond to segments. Visualization created with t-SNE.}
\label{fig:clustering}
\end{figure*}

\begin{figure*}

\subfloat[Psychiatry ]{
\includegraphics[width=0.45\linewidth]{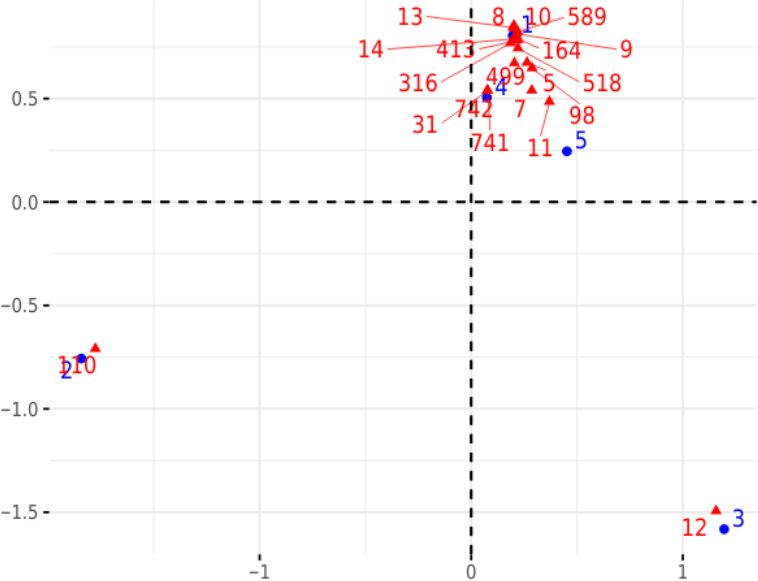}}
\subfloat[Family medicine]{
\includegraphics[width=0.45\linewidth]{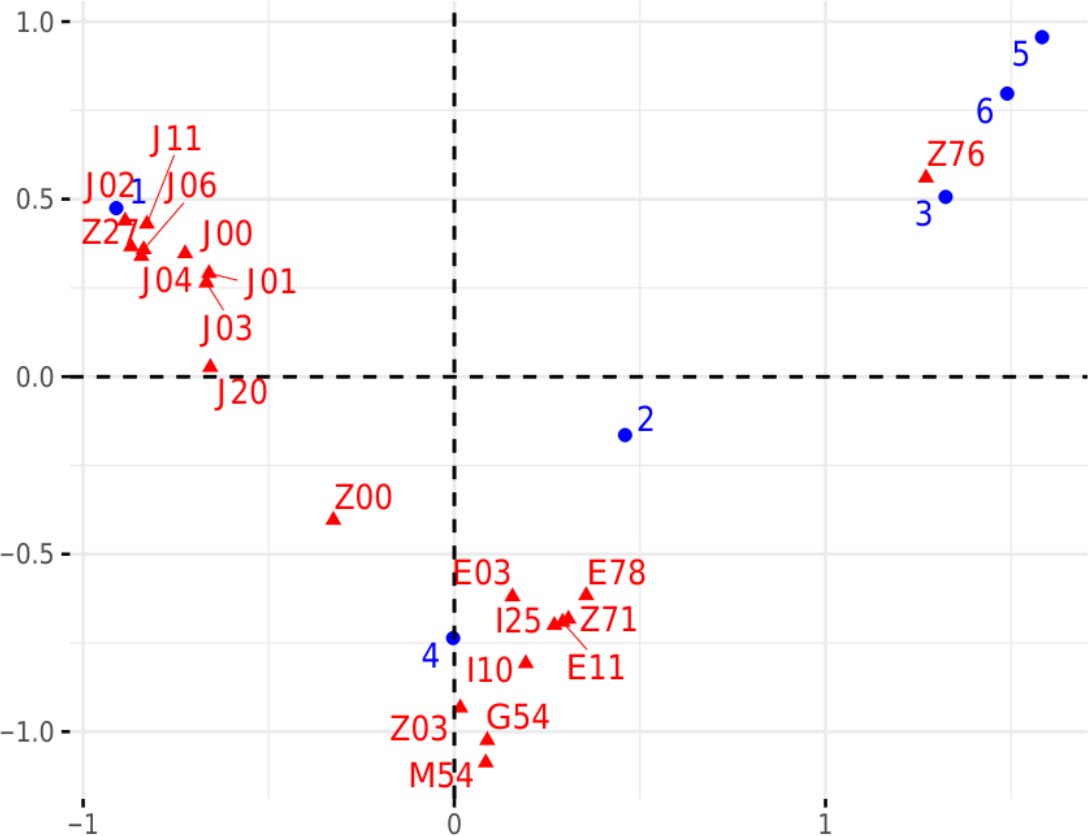}}
\caption{Correspondence analysis  between clusters and doctors' IDs for psychiatry clustering (panel a) and  between clusters and ICD-10 codes for family medicine clustering (panel b). Clusters 2 and 3 in panel a are perfectly fitted to a single doctor. }
\label{fig:doctors}


\end{figure*}

For computing visit embeddings we chose embeddings of dimensionality 20, since this resulted in the best accuracy of the most restrictive analogy task and it allowed us to perform more efficient computations than higher dimensional representations.
Figure~\ref{fig:analogies} illustrates PCA projection of term embeddings from four categories of analogies.

\subsection{Visits Clustering}

Clustering was performed separately for each specialty of doctors. 
Figure~\ref{fig:clustering} illustrates two-dimensional t-SNE projections of visit embeddings coloured by clusters~\cite{maaten2008visualizing}. For some domains clusters are very clear and separated (Figure~\ref{cardiology}). This corresponds with the high stability of clustering measured by Rand index.

In order to validate the proposed methodology we evaluate how clear are derived segments when it comes to medical diagnoses (ICD-10). No information about recommendations nor diagnosis is used in the phase of clustering to prevent data leakage.  

Figure~\ref{fig:doctors} (b) shows correspondence analysis between clusters and ICD-10 codes for family medicine clustering. There appeared two large groups of codes: the first related to diseases of the respiratory system (J) and the second related to other diseases, mainly endocrine, nutritional and metabolic diseases (E) and diseases of the circulatory system (I). The first group corresponds to Cluster 1 and the second to Cluster 4. Clusters 3, 5 and 6 (the smallest clusters in this clustering) covered Z76 ICD-10 code (encounter for issue of repeat prescription).
We also examined the distribution of doctors' IDs in the obtained clusters. It turned out that some clusters covered almost exactly descriptions written by one doctor. This happen in the specialties where clusters are separated with large margins (e.g. psychiatry, pediatrics, cardiology). Figure~\ref{fig:doctors} (a) shows correspondence analysis between doctors' IDs and clusters for psychiatry clustering.

\subsection{Recommendations in Clusters}

According to the main goal of our clustering described in Introduction, we would like to obtain similar recommendations inside every cluster. Hence we examined the frequency of occurrence of the recommendation terms in particular clusters.

We examined terms of recommendations related to one of five categories: procedure to carry out by patient, examination, treatment, diet and medicament. Table~\ref{tab:recom} shows an example of an analysis of the most common recommendations in clusters in gynecology clustering. In order to find only characteristic terms for clusters we filtered the terms which belong to one of 15 the most common terms in at least three clusters.

\section{Conclusions and Applications}

We proposed a new method for clustering of visits in health centers based on descriptions written by doctors. We validated this new method on a new large corpus of Polish medical records. For this corpus we identified medical concepts and created their embeddings with GloVe algorithm. The quality of the embeddings was measured by the specific analogy task designed specifically for this corpus. It turns out that analogies work well, what ensures that concept embeddings store some useful information.

Clustering was performed on visits embedding created based on word embedding. Visual and numerical examination of derived clusters showed an interesting structure among visits. As we have shown obtained segments are linked with medical diagnosis even if the information about recommendations or diagnosis were not used for the clustering. This additionally convinces that the identified structure is related to some subgroups of medical conditions.

Obtained clustering can be used to assign new visits to already derived clusters. Based on descriptions of an interview or a description of patient examination we can identify similar visits and show corresponding recommendations. 

\section*{Acknowledgments}
This work was financially supported by NCBR Grant POIR.01.01.01-00-0328/17. PBi was supported by NCN Opus grant 2016/21/B/ST6/02176.

\bibliography{ismis2020}

\begin{thebibliography}{28}
\providecommand{\natexlab}[1]{#1}
\providecommand{\url}[1]{\texttt{#1}}
\expandafter\ifx\csname urlstyle\endcsname\relax
  \providecommand{\doi}[1]{doi: #1}\else
  \providecommand{\doi}{doi: \begingroup \urlstyle{rm}\Url}\fi

\bibitem[{Apostolova} et~al.(2009){Apostolova}, {Channin}, {Demner-Fushman},
  {Furst}, {Lytinen}, and {Raicu}]{5334831}
E.~{Apostolova}, D.~S. {Channin}, D.~{Demner-Fushman}, J.~{Furst},
  S.~{Lytinen}, and D.~{Raicu}.
\newblock Automatic segmentation of clinical texts.
\newblock In \emph{Proceedings of EMBC}, pages 5905--5908, 2009.

\bibitem[Banea et~al.(2014)Banea, Chen, Mihalcea, Cardie, and
  Wiebe]{banea2014simcompass}
C.~Banea, D.~Chen, R.~Mihalcea, C.~Cardie, and J.~Wiebe.
\newblock Simcompass: Using deep learning word embeddings to assess cross-level
  similarity.
\newblock In \emph{Proceedings of SemEval}, pages 560--565, 2014.

\bibitem[Biecek(2018)]{DALEX}
P.~Biecek.
\newblock {DALEX}: {E}xplainers for {C}omplex {P}redictive {M}odels in {R}.
\newblock \emph{Journal of Machine Learning Research}, 19\penalty0
  (84):\penalty0 1--5, 2018.

\bibitem[Bodenreider(2004)]{bodenreider2004unified}
O.~Bodenreider.
\newblock The unified medical language system ({UMLS}): integrating biomedical
  terminology.
\newblock \emph{Nucleic acids research}, 32\penalty0 (suppl\_1):\penalty0
  D267--D270, 2004.

\bibitem[Chiu et~al.(2016)Chiu, Crichton, Korhonen, and Pyysalo]{chiu2016train}
B.~Chiu, G.~Crichton, A.~Korhonen, and S.~Pyysalo.
\newblock How to train good word embeddings for biomedical {NLP}.
\newblock In \emph{Proceedings of BioNLP}, pages 166--174, 2016.

\bibitem[Choi et~al.(2016{\natexlab{a}})Choi, Bahadori, Searles, Coffey,
  Thompson, Bost, Tejedor-Sojo, and Sun]{choi2016multi}
E.~Choi, M.~T. Bahadori, E.~Searles, C.~Coffey, M.~Thompson, J.~Bost,
  J.~Tejedor-Sojo, and J.~Sun.
\newblock Multi-layer representation learning for medical concepts.
\newblock In \emph{SIGKDD Proceedings}, pages 1495--1504. ACM,
  2016{\natexlab{a}}.

\bibitem[Choi et~al.(2016{\natexlab{b}})Choi, Schuetz, Stewart, and
  Sun]{choi2016medical}
E.~Choi, A.~Schuetz, W.~F. Stewart, and J.~Sun.
\newblock Medical concept representation learning from electronic health
  records and its application on heart failure prediction.
\newblock \emph{arXiv preprint arXiv:1602.03686}, 2016{\natexlab{b}}.

\bibitem[Choi et~al.(2016{\natexlab{c}})Choi, Chiu, and
  Sontag]{choi2016learning}
Y.~Choi, C.~Y.-I. Chiu, and D.~Sontag.
\newblock Learning low-dimensional representations of medical concepts.
\newblock \emph{AMIA Summits on Translational Science}, page~41,
  2016{\natexlab{c}}.

\bibitem[De~Boom et~al.(2016)De~Boom, Van~Canneyt, Demeester, and
  Dhoedt]{de2016representation}
C.~De~Boom, S.~Van~Canneyt, T.~Demeester, and B.~Dhoedt.
\newblock Representation learning for very short texts using weighted word
  embedding aggregation.
\newblock \emph{Pattern Recognition Letters}, 80:\penalty0 150--156, 2016.

\bibitem[De~Vine et~al.(2014)De~Vine, Zuccon, Koopman, Sitbon, and
  Bruza]{de2014medical}
L.~De~Vine, G.~Zuccon, B.~Koopman, L.~Sitbon, and P.~Bruza.
\newblock Medical semantic similarity with a neural language model.
\newblock In \emph{Proceedings of CIKM}, pages 1819--1822. ACM, 2014.

\bibitem[Fetter et~al.(1980)Fetter, Shin, Freeman, Averill, and
  Thompson]{casemix}
R.~B. Fetter, Y.~Shin, J.~L. Freeman, R.~F. Averill, and J.~D. Thompson.
\newblock Case mix definition by diagnosis-related groups.
\newblock \emph{Medical care}, 18\penalty0 (2):\penalty0 i--53, 1980.

\bibitem[Frantzi et~al.(2000)Frantzi, Ananiadou, and Mima]{fran:etal}
K.~Frantzi, S.~Ananiadou, and H.~Mima.
\newblock Automatic {R}ecognition of {M}ulti-{W}ord {T}erms: the
  {C}-value/{NC}-value {M}ethod.
\newblock \emph{Int. Journal on Digital Libraries}, 3:\penalty0 115--130, 2000.

\bibitem[{Ganesan} and {Subotin}(2014)]{7004390}
K.~{Ganesan} and M.~{Subotin}.
\newblock A general supervised approach to segmentation of clinical texts.
\newblock In \emph{IEEE International Conference on Big Data}, pages 33--40,
  2014.

\bibitem[Gordon et~al.(2019)Gordon, Grantcharov, and Rudzicz]{Gordon}
L.~Gordon, T.~Grantcharov, and F.~Rudzicz.
\newblock {Explainable Artificial Intelligence for Safe Intraoperative Decision
  Support}.
\newblock \emph{JAMA Surgery}, 154\penalty0 (11):\penalty0 1064--1065, 11 2019.
\newblock ISSN 2168-6254.

\bibitem[Jaworski and Kozakoszczak(2016)]{jaworski2016eniam}
W.~Jaworski and J.~Kozakoszczak.
\newblock {ENIAM}: Categorial syntactic-semantic parser for {P}olish.
\newblock In \emph{Proceedings of {COLING}}, pages 243--247, 2016.

\bibitem[Jaworski et~al.(2018)Jaworski, Oklesi{\'n}ski, Lupa, Rutkowski,
  Kozakoszczak, Przetacka, Tele{\.z}y{\'n}ska, Antonowicz, Markiewicz,
  Kowalewski, Pie{\'n}kosz, and Morusiewicz]{11321/538}
W.~Jaworski, D.~Oklesi{\'n}ski, J.~Lupa, S.~Rutkowski, J.~Kozakoszczak,
  J.~Przetacka, H.~Tele{\.z}y{\'n}ska, B.~Antonowicz, A.~Markiewicz,
  J.~Kowalewski, M.~Pie{\'n}kosz, and A.~Morusiewicz.
\newblock Categorial parser, 2018.
\newblock {CLARIN}-{PL} digital repository.

\bibitem[Kobyli{\'{n}}ska et~al.(2019)Kobyli{\'{n}}ska, Miko{\l}ajczyk, Adamek,
  Or{\l}owski, and Biecek]{Kobylinska}
K.~Kobyli{\'{n}}ska, T.~Miko{\l}ajczyk, M.~Adamek, T.~Or{\l}owski, and
  P.~Biecek.
\newblock Explainable machine learning for modeling of early postoperative
  mortality in lung cancer.
\newblock In \emph{Artificial Intelligence in Medicine: Knowledge
  Representation and Transparent and Explainable Systems}, pages 161--174.
  Springer, 2019.

\bibitem[Maaten and Hinton(2008)]{maaten2008visualizing}
L.~v.~d. Maaten and G.~Hinton.
\newblock Visualizing data using {t-SNE}.
\newblock \emph{Journal of machine learning research}, 9\penalty0
  (Nov):\penalty0 2579--2605, 2008.

\bibitem[Marciniak et~al.(2016)Marciniak, Mykowiecka, and
  Rychlik]{mar:myk:rych:lrec16}
M.~Marciniak, A.~Mykowiecka, and P.~Rychlik.
\newblock {TermoPL} --- a flexible tool for terminology extraction.
\newblock In \emph{Proceedings of {LREC}}, pages 2278--2284, Portorož,
  Slovenia, 2016. ELRA.

\bibitem[Mikolov et~al.(2013)Mikolov, Chen, Corrado, and
  Dean]{mikolov2013efficient}
T.~Mikolov, K.~Chen, G.~Corrado, and J.~Dean.
\newblock Efficient estimation of word representations in vector space.
\newblock \emph{arXiv preprint arXiv:1301.3781}, 2013.

\bibitem[Minarro-Gim{\'e}nez et~al.(2014)Minarro-Gim{\'e}nez, Marin-Alonso, and
  Samwald]{minarro2014exploring}
J.~A. Minarro-Gim{\'e}nez, O.~Marin-Alonso, and M.~Samwald.
\newblock Exploring the application of deep learning techniques on medical text
  corpora.
\newblock \emph{Studies in Health Technology and Informatics}, 205:\penalty0
  584--588, 2014.

\bibitem[Newman-Griffis et~al.(2017)Newman-Griffis, Lai, and
  Fosler-Lussier]{newman2017insights}
D.~Newman-Griffis, A.~M. Lai, and E.~Fosler-Lussier.
\newblock Insights into analogy completion from the biomedical domain.
\newblock \emph{arXiv preprint arXiv:1706.02241}, 2017.

\bibitem[Orosz et~al.(2013)Orosz, Nov{\'a}k, and
  Pr{\'o}sz{\'e}ky]{hungariansegmentation}
G.~Orosz, A.~Nov{\'a}k, and G.~Pr{\'o}sz{\'e}ky.
\newblock Hybrid {T}ext {S}egmentation for {H}ungarian {C}linical {R}ecords.
\newblock In \emph{Proceedings of MICAI}, pages 306--317. Springer, 2013.

\bibitem[Pennington et~al.(2014)Pennington, Socher, and
  Manning]{pennington2014glove}
J.~Pennington, R.~Socher, and C.~Manning.
\newblock Glove: Global vectors for word representation.
\newblock In \emph{Proceedings of EMNLP}, pages 1532--1543, 2014.

\bibitem[Rand(1971)]{rand1971objective}
W.~M. Rand.
\newblock Objective criteria for the evaluation of clustering methods.
\newblock \emph{Journal of the American Statistical association}, 66\penalty0
  (336):\penalty0 846--850, 1971.

\bibitem[Ruffini et~al.(2017)Ruffini, Gavald{\`a}, and
  Lim{\'o}n]{ruffini2017clustering}
M.~Ruffini, R.~Gavald{\`a}, and E.~Lim{\'o}n.
\newblock Clustering patients with tensor decomposition.
\newblock \emph{arXiv preprint arXiv:1708.08994}, 2017.

\bibitem[Ward~Jr(1963)]{ward1963hierarchical}
J.~H. Ward~Jr.
\newblock Hierarchical grouping to optimize an objective function.
\newblock \emph{Journal of the American statistical association}, 58\penalty0
  (301):\penalty0 236--244, 1963.

\bibitem[Waszczuk(2012)]{wasz:12}
J.~Waszczuk.
\newblock Harnessing the {CRF} complexity with domain-specific constraints.
  {T}he case of morphosyntactic tagging of a highly inflected language.
\newblock In \emph{Proceedings of {COLING}}, pages 2789--2804, 2012.

\end{thebibliography}

\end{document}